\pdfoutput=1
\let\accentvec\vec
\documentclass[runningheads]{llncs}

\let\vec\accentvec
\usepackage{cite}
\usepackage{marvosym}
\usepackage{amsmath,amssymb,amsfonts}
\usepackage[noend]{algorithmic}
\usepackage{graphicx}
\usepackage{textcomp}
\usepackage{appendix}
\usepackage{xcolor}
\usepackage{pifont}
\usepackage{wrapfig}
\usepackage{url}
\usepackage{subfigure}
\usepackage{algorithm}  
\usepackage{algorithmic,eqparbox,array}
\usepackage{amsfonts,amssymb}
\usepackage{indentfirst}
\usepackage{verbatim}
\usepackage{float}
\usepackage{color}
\usepackage{cite}
\usepackage{threeparttable}
\def\BibTeX{{\rm B\kern-.05em{\sc i\kern-.025em b}\kern-.08em
		T\kern-.1667em\lower.7ex\hbox{E}\kern-.125emX}}

\usepackage{tikz}
\usetikzlibrary{shapes.geometric, arrows}

\usepackage{graphicx}

\begin{document}
	\title{
	FedSel: Federated SGD under Local Differential Privacy with Top-k Dimension Selection
	}
	\titlerunning{FedSel: Federated SGD under LDP with Top-k Dimension Selection}

	\author{Ruixuan Liu\inst{1}, Yang Cao\inst{2}, Masatoshi Yoshikawa\inst{2}, Hong Chen\inst{1}\textsuperscript{(\Letter)} }
	\authorrunning{R. Liu et al.}

	\institute{
		\textsuperscript{1}Renmin University of China, Beijing, China\\
		\email{\{ruixuan.liu, chong\}@ruc.edu.cn\\}
		\textsuperscript{2}Kyoto University, Kyoto, Japan\\
		\email{\{yang, yoshikawa\}@i.kyoto-u.ac.jp}
	}
	
	\maketitle
	
	\begin{abstract}
		As massive data are produced from small gadgets, federated learning on mobile devices has become an emerging trend.
		In the federated setting, Stochastic Gradient Descent (SGD) has been widely used in federated learning for various machine learning models.
		To prevent privacy leakages from gradients that are calculated on users' sensitive data, local differential privacy (LDP) has been considered as a privacy guarantee in federated SGD recently.
		However, the existing solutions have a dimension dependency problem: the injected noise is substantially proportional to the dimension $d$.
		In this work, we propose a two-stage framework \textit{FedSel} for federated SGD under LDP to relieve this problem. 
		Our key idea is that not all dimensions are equally important so that we privately select Top-k dimensions according to their contributions in each iteration of federated SGD.
		Specifically, we propose three private dimension selection mechanisms and adapt the gradient accumulation technique to stabilize the learning process with noisy updates.
		We also theoretically analyze privacy, accuracy and time complexity of \textit{FedSel}, which outperforms the state-of-the-art solutions.
		Experiments on real-world and synthetic datasets verify the effectiveness and efficiency of our framework.
	\end{abstract}
	\section{Introduction}
	\noindent
	Nowadays, massive peripheral gadgets, such as mobile phones and wearable devices, produce an enormous volume of personal data, which boosts the process of Federated Learning (FL). 
	In the cardinal FL setting, Stochastic Gradient Descent (SGD) has been widely used \cite{mcmahan2017communication, bonawitz2019towards, yang2019federated, mcmahan2016Federated} for various machine learning models, such as Logistic Regression, Support Vector Machine and Neural Networks.
    The server iteratively updates a global model for $E$ epochs based on gradients of the objective function, which are collected from batches of $m$ clients' $d$-dimensional local updates.
	Nevertheless, users' data are still under threats of privacy attacks\cite{zhu2019deep,nasr2019comprehensive,Fredrikson2015Model,wang2019beyond} if the raw gradients are transmitted to an untrusted server.
	Local differential privacy (LDP) provides a rigorous guarantee to perturb users' sensitive data before sending to an untrusted server.
	Many LDP mechanisms have been proposed for different computational tasks or data types, such as matrix factorization \cite{shin2018Privacy}, key-valued data \cite{gu2019pckv,ye2019privkv} and multidimensional data \cite{nguyen2016collecting, wang2019collecting, duchi2018Minimax, gu_supporting_2019, gu2019providing}.
	
	However, applying LDP to federated SGD faces a nontrivial challenge when the dimension $d$ is large.
	First, although some LDP mechanisms \cite{nguyen2016collecting, wang2019collecting, duchi2018Minimax} proposed for multidimensional data are shown to be applicable to SGD (i.e., flat solution in Table \ref{tab-contribution}), the injected noise is substantially proportional to the dimension $d$. 
	Besides, in order to obtain an acceptable accuracy, the required batch size of clients (i.e., $m$) linearly depends on $d$.
	As the clients in federated learning have the full autonomy for their local data and can decide when and how to join the training process, a large required batch size impedes the model's applicability in practice.
	Second, a recent work \cite{shin2018Privacy} (i.e., compressed solution in Table \ref{tab-contribution}) attempts to solve this problem by reducing the dimension from $d$ to $q$ with random projection \cite{william1984extensions}, which is a well-studied dimension reduction technique.
	However, as this method randomly discards some dimensions, it introduces a large recovery error which may damage the learning performance.
	
	Our idea is that not all the dimensions are equally ``important'' so that we privately select Top-k dimensions according to their contributions (i.e., the absolute values of gradients) in each iteration of SGD. 
	A simple method for private Top-1 selection is to employ exponential mechanism \cite{dwork2014the(book)} that returns a private dimension with a probability proportional to its absolute value of gradient.
	However, the challenge is how to design Top-k selection mechanisms under LDP for federated SGD with better selection strategies.
	Besides the absence of such private Top-k selection mechanisms, another challenge is that discarding delayed gradients causes the convergence issues, especially in our private setting where extra noises are injected.
	Although we can accumulate delayed gradients with momentum\cite{sun2019sparse}, it still requires additional design to stabilize the learning process with noisy updates.
	
	\begin{table}[t]
		\caption{Overview Comparison (the number of epochs $E$, the dimension of a gradient vector $d$, the compressed dimension $q$, the privacy budget for training $\epsilon$, the amount of participants in one aggregation $m$. See detailed analyses in Sects.\ref{subsec-ourline} and \ref{subsec-time-acc}).}\label{tab-contribution}
		\begin{center}
			\vspace{-8pt}
			\begin{tabular}{ccc}
				\hline
				LDP-SGD solutions & upper bound of noise & ~~~~lower bound of batch size\\
				\hline
				flat solution\cite{nguyen2016collecting, wang2019collecting, duchi2018Minimax} & $O(\frac{E \sqrt{d\log d}}{\epsilon\sqrt{m}})$ & $\Omega(\frac{ E^2 d \log d}{\epsilon^2})$\\ 
				compressed solution\cite{shin2018Privacy} & $O(\frac{E  \sqrt{q\log q}}{\epsilon\sqrt{m}})$ & $\Omega(\frac{E^2 q \log q}{\epsilon^2})$\\
				our solution & $O(\frac{E \sqrt{\log d}}{\epsilon\sqrt{md}})$ & $\Omega(\frac{E^2 \log d}{d\epsilon^2})$\\
				\hline
			\end{tabular}
		\end{center}
	\vspace{-27pt}
	\end{table}
	
	\textbf{Contributions:}
	In this work, we take the first attempt to mitigate the dimension dependency problem in federated learning with LDP. 
	We design, implement and evaluate a two-stage $\epsilon$-LDP framework for federated SGD. 
	As shown in Table \ref{tab-contribution}, comparing to the state-of-the-art techniques, our framework relieves the effect of the number of dimensions $d$ on the injected noise and the required batch size (i.e., size of clients in each iteration; the lower, the better).
	Our contributions are summarized below.
	
	\begin{itemize}
		\item First, we propose a two-stage framework \textit{FedSel} with  \textit{Dimension Selection} and \textit{Value Perturbation}. With our privacy analysis in Sect.3.2, this framework satisfies $\epsilon$-LDP for any client's local vector. Our theoretical utility analysis in Sect.3.3 shows that it significantly reduces the dimensional dependency on LDP estimation error from $O(\sqrt{d})$ to $O(1/\sqrt{d})$.
		We also enhance the framework to avoid the loss of accuracy by accumulating delayed gradients. 
		Intuitively, delayed gradients can improve the empirical performance and fix the convergence issues. In order to further stabilize the learning in the private setting with noises injected, we modify an existing accumulation\cite{sun2019sparse}. Our analysis and experiments validate that this modification reduces the variance of noisy updates. (Section 3)
		\item Second, we instantiate the \textit{Dimension Selection} stage with three mechanisms, which are general and independent of the second stage of value perturbation. The privacy guarantee for the selection is provided. Besides, we show the advance of utility and computation cost by extending Top-1 to Top-k case with analysis and experiments. (Section 4)
		\item Finally, we perform extensive experiments on synthetic and real-world datasets to evaluate the proposed framework and the private Top-k dimension selection mechanisms. We also implement a \textit{hyper-parameters-free} strategy to automatically allocate the privacy budgets between the two stages with better utility. Significant improvements are shown in test accuracy comparing with the existing solutions\cite{nguyen2016collecting,wang2019collecting, duchi2018Minimax, shin2018Privacy}. (Section 5)
	\end{itemize}
	
	The remainder of this paper is organized as follows. Section \ref{sec-sec-prelimina} presents the technical background. Section \ref{sec-overview} illustrates the two-stage privatized framework with analyses. Section \ref{sec-selection} proposes the \underline{Exp}onential Mechanism (EXP) for Top-1 selection and extends to Top-k case with \underline{P}erturbed \underline{E}ncoding Mechanism (PE) and \underline{P}erturbed \underline{S}ampling Mechanism (PS). Section \ref{sec-experiments} provides results on both synthetic and real-world datasets and a \textit{hyper-parameters-free} strategy. Section \ref{sec-related works} gives an overview of related works. Section \ref{sec-conclusions} concludes the paper.
	
	\section{Preliminaries}\label{sec-sec-prelimina}
	\subsection{Federated SGD}
	\noindent
	Suppose a learning task defines the objective loss function $L(w;x,y)$ on example $(x,y)$ with parameters $w\in \mathbb{R}^d$.
	The goal of learning is to construct an empirical minimization as $w^* = \arg\min_w \frac{1}{N}\sum_{i=1}^N L(w;x_i,y_i)$ over $N$ clients' data.
	For a single iteration, a batch of $m$ clients updates local models in parallel with the distributed global parameters.
	Then they transmit local model updates to the server for an average mean gradient to update the global model as:
	$
	\setlength{\abovedisplayskip}{3pt}
	\setlength{\belowdisplayskip}{3pt}
	w^t \leftarrow w^{t-1} - \alpha \frac{1}{m} \sum_{i=1}^{m} \bigtriangledown L(w^{t-1};x_i,y_i).
	$
	Without loss of generality, we describe our framework with the classic setting with one local update for each round.
	
	\subsection{Local Differential Privacy}
	\noindent
	Local differential privacy (LDP) is proposed for collecting sensitive data through local perturbations without any assumption on a trusted server.
	$\mathcal{M}$ is a randomized algorithm that takes a vector $v$ as input and outputs a perturbed vector $v^*$. $\epsilon$-LDP is defined on $\mathcal{M}$ with a privacy budget $\epsilon$ as follows.
	\begin{definition}[Local Differential Privacy\cite{duchi2013local}]\label{def-ldp}
		A randomized algorithm $\mathcal{M}$ satisfies $\epsilon$-LDP if and only if the following is true for any two possible inputs $v, v' \in \mathcal{V}$ and output $v^*$:
		$
		\setlength{\abovedisplayskip}{3pt}
		\setlength{\belowdisplayskip}{3pt}
		Pr[\mathcal{M}(v)=v^*] \le e^\epsilon \cdot Pr[\mathcal{M}(v')=v^*].
		$
	\end{definition}
	
	\subsection{Problem Definition}
	\noindent
	This paper studies the problem of federated SGD with LDP. 
	Note that in the practical non-private setting, the original gradient $g^t \leftarrow \bigtriangledown L(w^{t-1};x,y)$ can be sparsified\cite{alistarh2018convergence} or quantized\cite{alistarh2017qsgd} before a transmission.
	For generality, we do not limit the form of local gradient and use $v_i$ to denote the local gradient calculated from client $u_i$'s record $(x_i, y_i)$ and the global parameters $w^{t-1}$. 
	
	Suppose the global model iterates for $E$ epochs with a learning rate $\alpha$ and a total privacy budget $\epsilon$ for each client.
	For a single epoch, clients are partitioned into batches with size $m$. 
	Then the privatized mechanism $\mathcal{M}$ privatizes $m$ local updates before they are aggregated by the untrusted server for one iteration. 
	The global model is updated as:
	$w^t \leftarrow w^{t-1} -\alpha \frac{1}{m}\sum_{i=1}^m \mathcal{M}(v_i)$.	
	We aim to propose an $\epsilon$-LDP framework with a specialized mechanism $\mathcal{M}$ for private federated training against the untrusted server.
	Moreover, we attempt to mitigate the dimension dependency problem for a higher accuracy.
	
	\section{Two-stage LDP Framework: FedSel}\label{sec-overview}
\begin{figure}[!ht]
		\centering
		\vspace{-10pt}
		\includegraphics[width=0.99\textwidth,trim=8 5 20 60,clip]{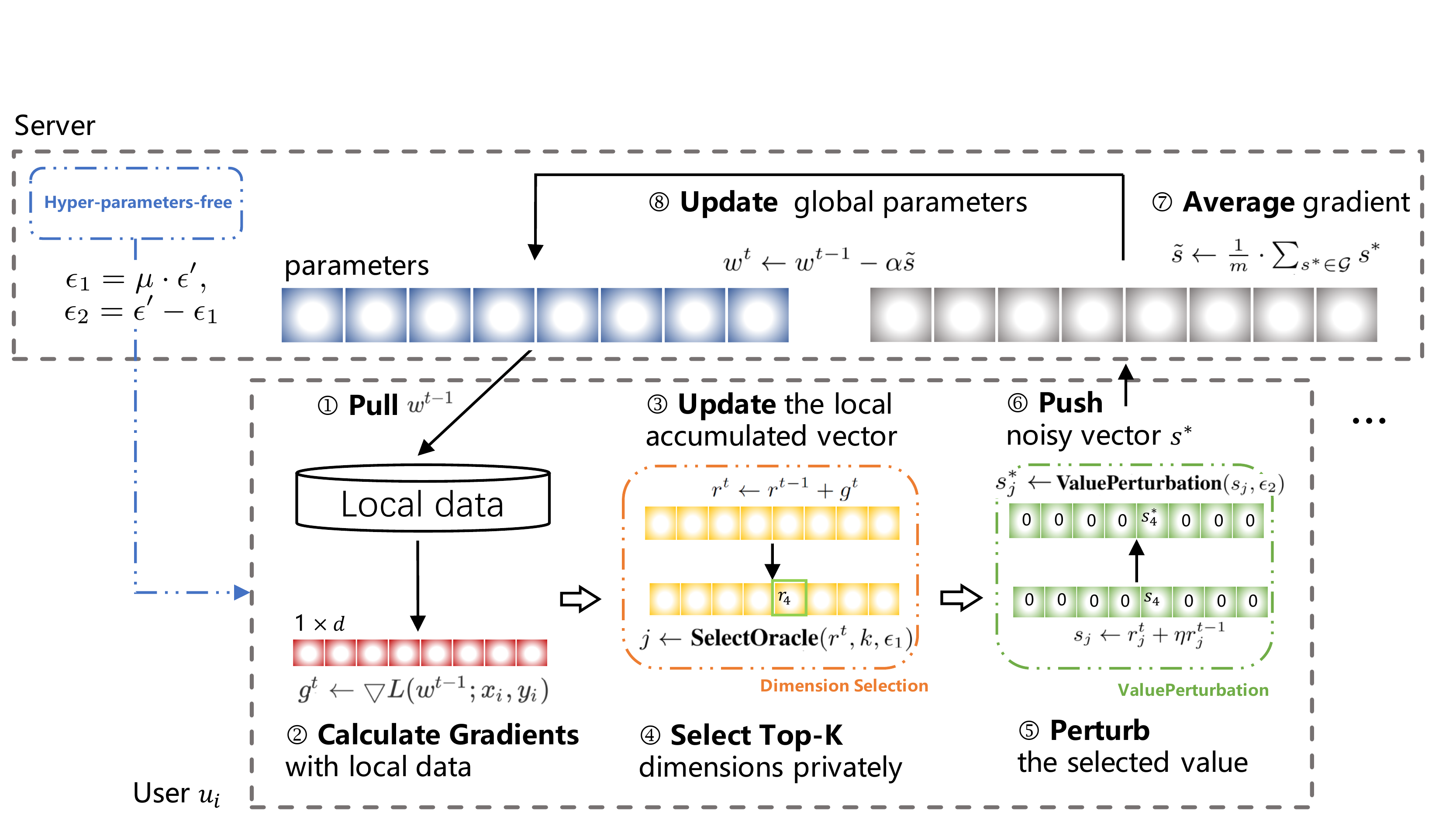} 
		\setlength{\abovedisplayskip}{-35pt}
		\caption{Two-stage LDP framework.}
		\label{overview-framework}
	\end{figure}

	\noindent
	In this section, we propose a two-stage framework \textit{FedSel} with dimension selection and value perturbation as shown in Fig.\ref{overview-framework}.
	The framework and differences from existing works are presented in Sect.\ref{subsec-ourline}.
	We prove the privacy guarantee in Sect.\ref{subsec-decompose} and analyze the stability improvement in Sect.\ref{subsec-adapted}.
	
	\begin{algorithm}[!ht]
		\caption{Two-stage LDP framework of federated learning}
		$E, N, m, \mu, \alpha, \epsilon$ for the server; $\epsilon_1, \epsilon_2, \eta, k$ for $N$ clients;
		\begin{algorithmic}[1]\label{alg-generalProtocol}
			\renewcommand{\algorithmicrequire}{ \textbf{GlobalUpdate:}}
			\renewcommand{\algorithmicensure}{ \textbf{LocalUpdate:}}
			\REQUIRE 
			\STATE initialize $t=1,w^0,r_i^0\leftarrow \{0\}^d$ for $i\in[N]$ 
			\STATE $\mu \leftarrow$ \textbf{HyperParametersFree($m,\epsilon,d$)}\label{line-hyper}
			\STATE initialize $\epsilon^{\prime}=\epsilon/E, \epsilon_1 = \mu \cdot\epsilon^{\prime}, \epsilon_2 = \epsilon^{\prime} - \epsilon_1$
			\FOR {each epoch $1,\cdots, E$}
			\FOR {each sample batch with size $m$}
			\STATE initialize $\mathcal{G}$ as an empty set
			\FOR {each client} 
			\STATE $s^*\leftarrow$ LocalUpdate($w^{t-1}, \epsilon_1, \epsilon_2, \eta, k$)
			\STATE add $s^*$ to $\mathcal{G}$
			\ENDFOR
			\STATE $\tilde{s} \leftarrow \frac{1}{m} \cdot \sum_{s^*\in \mathcal{G}} s^*$  \COMMENT{aggregation}
			\STATE $w^t \leftarrow w^{t-1} - \alpha \tilde{s}$, $t = t+1$\label{line-scale}
			\ENDFOR
			\ENDFOR
			\RETURN final global model $w$
			
			\ENSURE($w^{t-1}, \epsilon_1, \epsilon_2, \eta, k$)
			\STATE initialize $s^* \leftarrow \{0\}^d$
			\STATE $g^t \leftarrow \bigtriangledown L(w^{t-1};x,y)$\label{line-gradient}
			\STATE $r^t \leftarrow r^{t-1} + g^t$ \COMMENT{adapted accumulation}\label{line-acc}
			\STATE $j \leftarrow $ \textbf{SelectOracle}$(r^t, k, \epsilon_1)$\label{line-selectOracle}
			\STATE $s_j\leftarrow r^t_j + \eta r_j^{t-1}$\label{line-momentum}
			\STATE $s_j^* \leftarrow \textbf{ValuePerturbation}(s_j, \epsilon_2 ) ,  r^t_j \leftarrow 0$\label{line-valuePerturbation}
			\RETURN  $s^*$
		\end{algorithmic}
	\end{algorithm}
	
	\subsection{Overview}\label{subsec-ourline}
	\noindent
	We now illustrate the proposed framework in Algorithm \ref{alg-generalProtocol} and compare it with the flat solutions\cite{wang2019collecting, nguyen2016collecting, duchi2018Minimax} and the compressed solution\cite{shin2018Privacy}.
	In our framework, the server first initiates the ratio $\mu\in[0, 1]$ for privacy budget allocation and starts the iteration.
	For the procedure on a local device:
	(i) Current gradient $g^t$ is accumulated with previously delayed gradients $r^{t-1}$ (line \ref{line-acc}). (ii) An important dimension index is privately selected by \textbf{Dimension Selection} (line \ref{line-selectOracle}). (iii) The value of the selected dimension plus its momentum (line \ref{line-momentum}) is perturbed by \textbf{Value Perturbation} (line \ref{line-valuePerturbation}). 
	The accumulation in step(i) and the momentum in step(iii) derive from the work of Sun et al.\cite{sun2019sparse} to compress local gradient with little loss of accuracy and memory cost. 
	We adapt the accumulation in our framework to stabilize the iteration with noises.
	The step(ii) is designed to alleviate the dimensional bottleneck and we analyze the improvement of accuracy in Sect.\ref{subsec-time-acc}.
	 
	\emph{Comparison with existing works.}
	The most significant difference from existing works is the way of deciding which dimension to upload. 
	In the flat solution, each client randomly samples and perturbs $c$ dimensions from $d$. 
	the perturbed value is enlarged by $d/c$ for an unbiased mean estimation. Thus, the injected noise is also amplified.
	For the compressed solution, $d$ dimensions of the gradient $g$ is reduced to $q$ dimensions by multiplying the vector $g$ with a public random matrix $\Phi_{d\times q}$ drawn from the Gaussian distribution with mean 0 and variance $1/q$.
	Then an index is randomly sampled from $q$ dimensions with its value perturbed. The estimated mean of compressed vector is then approximately recovered with the pseudo-inverse of $\Phi$. Even if the random projection has the strict isometry property, it ignores the meaning of the gradient magnitude and brings recovery error. Our framework differs the compressed solution because we utilize the magnitude property for gradient value.	
	
	\vspace{-5pt}
	\subsection{Privacy Guarantee}\label{subsec-decompose}
	\noindent
	In Algorithm \ref{alg-generalProtocol}, both the local accumulated vector $r$ and the vector with momentum $s$ are true local vectors that are directly calculated from the private data. So we abuse $v\in\mathbb{R}^d$ to denote them in the following statement.
	As previously defined by Shokri et al.\cite{shokri2015privacy}, there are two sources of information that we intend to preserve for the local vector: (i) how a dimension is selected and (ii) the value of the selected dimension.
	Let $z=\{0,1\}^d$ indicate the ground-truth Top-k status.
	We decompose the protection goal into following privacy definitions for dimension selection and value perturbation. 
	When combining the two stages together, we can provide an $\epsilon$-LDP guarantee with Theorem \ref{theo-e1e2}.
	\begin{definition}[LDP Dimension Selection]
		A randomized dimension selection algorithm $\mathcal{M}_1$ satisfies $\epsilon_1-$LDP if and only if for any two status vectors $z, z'\in\{0,1\}^d$ and any output $j\in[d]$:
		$
		Pr[\mathcal{M}_1(z)=j]\le e^{\epsilon_1} \cdot Pr[\mathcal{M}_1(z')=j].
		$
	\end{definition}
	\begin{definition}[LDP Value Perturbation] 
		A randomized value perturbation algorithm $\mathcal{M}_2$ satisfies $\epsilon_2-$LDP if and only if for any two numeric values $v_j,v'_j$ and any output $v_j^*$:
		\setlength{\abovedisplayskip}{3pt}
		\setlength{\belowdisplayskip}{3pt}
		$
		Pr[\mathcal{M}_2(v_j)=v_j^*]\le e^{\epsilon_2} \cdot Pr[\mathcal{M}_2(v'_j)=v_j^*].
		$
	\end{definition}
	\begin{theorem}\label{theo-e1e2}
		For a true local vector $v$, if the two-stage mechanism $\mathcal{M}$ first selects a dimension index $j$ with $\mathcal{M}_1$ and then perturbs $v_j$ with $\mathcal{M}_2$ under $\epsilon_1$-LDP and $\epsilon_2$-LDP respectively, $\mathcal{M}$ satisfies $(\epsilon_1+\epsilon_2)$-LDP.
	\end{theorem}
	\emph{proof.} 
	Theorem \ref{theo-e1e2} stands only when for any two possible local vectors $v,v'$, the conditional probabilities for $\mathcal{M}$ to give the same output $v^*$ satisfy the following condition:
	\setlength{\abovedisplayskip}{3pt}
	\setlength{\belowdisplayskip}{3pt}
	$
	Pr[v^*|v]\le e^{\epsilon_1+\epsilon_2} Pr[v^*|v'].
	$
	Let $z,z'$ denote selection status vectors of $v,v'$. 
	As we are considering the case where $v,v'$ have the same output $v^*$, we end the proof with:
	\begin{align*}
	\frac{Pr[v^*|v] }{Pr[v^*|v'] }
	= \frac{Pr[z|v]Pr[j|z]Pr[v_j^*|v_j]}{Pr[z'|v']Pr[j|z']Pr[v_j^*|v_j']} \le \frac{Pr[z|v]}{Pr[z'|v']}\cdot e^{\epsilon_1} e^{\epsilon_2}=e^{(\epsilon_1+\epsilon_2)}.
	\end{align*}

	\subsection{Variance Analysis for Accumulation}\label{subsec-adapted}
	\noindent
	In the existing non-private gradients accumulation\cite{sun2019sparse, lin2017deep, alistarh2018convergence, aji2017sparse}, local vectors are accumulated as $r^t=r^{t-1}+\alpha g^t$. 
	We adapt it to $r^t=r^{t-1}+g^t$ and scale it with the learning rate in line \ref{line-scale}. This aims to reduce the variance of noisy local updates and stabilize the iteration.
	Suppose the $j^{th}$ dimension of a gradient is selected after  $T$ rounds of delay, and $s_{i,j}$ denotes the gradient value. In each iteration of Algorithm \ref{alg-generalProtocol}, the update for the global parameter $w_j$ from user $u_i$ is denoted as $\bigtriangleup_T w_{i,j}=\frac{\alpha}{m} s_{i,j}^*$. If $\alpha g$ is accumulated\cite{sun2019sparse}, the update is $\bigtriangleup_T^\prime w_{i,j}=\frac{1}{m}(\alpha s_{i,j})^*$.
	
	For LDP mechanisms of mean estimation, the input range is $[-1, 1]$. Thus all inputs for a perturbation mechanism should be clipped to the defined input range. 
	When the value for the selected dimension is in the input range, we can easily have: $Var[\bigtriangleup_T w_{i,j}]=Var[\bigtriangleup_T^\prime w_{i,j}]$. It should be noted that our selected gradient value has significantly larger magnitude. When initial values of both cases are clipped to the input domain, we define the clipped input as $\xi$. Then we have: $Var[\bigtriangleup_T w_{i,j}]=\frac{\alpha^2}{m^2}Var[\xi^*], Var[\bigtriangleup_T^\prime w_{i,j}]=\frac{1}{m^2}Var[\xi^*]$.
	As the learning rate $\alpha$ is typically smaller than 1, such as 0.1, we then have Theorem \ref{theo-var-adaption}. Therefore, the slight adaption of scaling procedure can reduce the variance of single client's noisy update. The advance of a smaller variance will lead to a more accurate learning performance as validated in the experimental part, Fig.\ref{fig-adapted}.
	\begin{theorem} \label{theo-var-adaption}
		Accumulating $g$ instead of $\alpha g$ for the dimension $j$ of user $u_i$ has an update on the global model's parameter with a less variance: 
		\begin{align*}
		Var[\bigtriangleup_T w_{i,j}]\le Var[\bigtriangleup_T^\prime w_{i,j}], \text{ where } \alpha\in(0,1).
		\end{align*}
	\end{theorem}
	
	\section{Private Dimension Selection}\label{sec-selection}
	\noindent
	To instantiate the $\epsilon_1$-LDP dimension selection in Algorithm \ref{alg-generalProtocol} line \ref{line-selectOracle}, we design LDP mechanisms from Top-1 to Top-k case. Then, we analyze the accuracy and time complexity of proposed LDP mechanisms. 
	
	\subsection{Selection Mechanisms}
	
	\noindent
	\textbf{Exponential Mechanism (EXP): }
	Exponential mechanism\cite{dwork2014the(book)} is a natural building block for selecting private non-numeric data in the centralized differential privacy.
	We modify this classic method to meet the selection requirement.
	A client's accumulated vector $r$ is first sorted in ascending order of its absolute value.
	As a special case for Top-1 selection, the status vector in EXP is defined as: $z=\{1, \cdots, d\}^d$ instead of a binary vector.
	Intuitively, the dimension with the largest magnitude of its absolute value should be output with the highest probability.
	For the $j^{th}$ dimension, we assign the selection status as its ranking $z_j$.
	Thus, the index $j\in[d]$ is sampled unevenly with the probability $ \frac{\exp(\frac{\epsilon_1 z_j}{d-1})}{\sum_{i=1}^d \exp (\frac{\epsilon_1 z_i}{d-1})}$.
	The privacy guarantee is shown in Lemma \ref{exp_e1}.
	\begin{lemma}\label{exp_e1}
		EXP selection is $\epsilon_1-$locally differentially private.
	\end{lemma}
	\emph{proof.}
	Given any two possible ranking vectors as $z,z^{\prime} \in \{1, \cdots, d\}^d$. $j$ denotes any output index of EXP. 
	The following conditional probability ends the proof:
	$$
	\frac{Pr[j|z]}{Pr[j|z']} 
	=  \frac{\exp(\frac{\epsilon_1 z_j}{d-1})}{\sum_{i=1}^d \exp (\frac{\epsilon_1 z_i}{d-1})}/\frac{\exp(\frac{\epsilon_1 z_j^ \prime}{d-1})}{\sum_{i=1}^d \exp (\frac{\epsilon_1 z_i^{\prime}}{d-1})}
	\le \frac{\exp(\frac{\epsilon_1 \cdot d}{d-1})}{\exp(\frac{\epsilon_1\cdot 1}{d-1})}
	= e^\frac{\epsilon_1(d-1)}{d-1} =e^{\epsilon_1}.
	$$
	
	In order to fit various learning tasks, $k$ should be tunable. Thus, we propose two private Top-k methods to better control the selection.
	
	\vspace{5pt}
	\noindent
	\textbf{Perturbed Encoding Mechanism (PE): }
	The sorting step for the vector of absolute values $|r|$ in PE is the same as in EXP. Besides, a binary Top-k status vector $z$ is derived. Then we perturb the vector $z$ with the randomized response. Specifically, each status has a large probability $p$ to retain its value and a small probability $1-p$ to flip.
	For the privacy guarantee, $p=\frac{e^{\epsilon_1}}{e^{\epsilon_1}+1}$.
	Let $\acute{z}$ denote the privatized status vector.
	Since indices of non-zero elements in $\acute{z}$ are more likely to be Top-k dimensions, 
	we gather these elements as the sample set $\mathbb{S}$. 
	If $\mathbb{S}$ is empty, the client uploads $\bot$ and the server regards it as receiving a zero vector $s^*=\{0\}^d$. Elsewise the client randomly samples one dimension index from $\mathbb{S}$. The privacy guarantee is shown in Lemma \ref{lem-de_e1}.
	
	\begin{lemma}\label{lem-de_e1}
		PE selection is $\epsilon_1-$locally differentially private.
	\end{lemma}
	\emph{proof.}
	$\acute{z}$ denotes the perturbed status vector.
	The expected sparsity of $\acute{z}$ is:
	\begin{align*}
	l =& \mathbb{E}[||\acute{z}||_0]
	= \sum_{j=1}^{d}Pr[\acute{z}_j=1|z_j]
	= \sum_{j=1}^{k}Pr[\acute{z}_j=1|z_j=1] + \sum_{j=k+1}^{d}Pr[\acute{z}_j=1|z_j=0]\\
	=& k \cdot p + (d-k)\cdot(1-p),~\text{where  } p=\frac{e^{\epsilon_1}}{e^{\epsilon_1}+1}.
	\end{align*}
	Given any two possible selection status vectors as $z,z'\in\{0,1\}^d$ with $k$ non-zero elements, there are two cases for the output j:(i) If the sample set $\mathbb{S}$ is not empty, $j\in\mathbb{S}$. (ii) If the sample set $\mathbb{S}$ is empty, $j=\bot$.
	For the first case, we have:
	\begin{align*}
	\frac{Pr[j|z]}{Pr[j|z']} = \frac{\frac{1}{l}Pr[\acute{z}_j=1|z]}{\frac{1}{l}Pr[\acute{z}^{\prime}_j=1|z']}
	\le \frac{\frac{1}{l}Pr[\acute{z}_j=1|z_j=1]}{\frac{1}{l}Pr[\acute{z}_j^{\prime}=1|z_j'=0]}
	= \frac{e^{\epsilon_1}}{e^{\epsilon_1}+1}/\frac{1}{e^{\epsilon_1}+1}=e^{\epsilon_1}.
	\end{align*}
	For the second case, we end the proof with the conditional probability:
	\begin{align*}
	\frac{Pr[j=\bot|z]}{Pr[j=\bot|z']}
	= \frac{{(1-p)}^k\cdot p^{d-k}}{{(1-p)}^k\cdot p^{d-k}}=1
	\le& e^{\epsilon_1} \text{  (for }\epsilon_1\ge 0)
	\end{align*}
	
	\noindent
	\textbf{Perturbed Sampling Mechanism (PS): } PS selection has the same criterion as PE that regards Top-k as important dimensions. Intuitively, we define a higher probability $p$ to sample an index $j$ from the Top-k indices set $\{j\in[d]|{z}_j=1\}$ and elsewise, sample an index $j$ from non-Top-k dimensions $\{j\in[d]|{z}_j=0\}$ with a smaller probability $1-p$. With the privacy guarantee in Lemma \ref{lem-privacy-ps}, we define $p=\frac{e^{\epsilon_1}\cdot k}{d-k+e^{\epsilon_1}\cdot k}$.
	
	\begin{lemma}\label{lem-privacy-ps}
		PS selection is $\epsilon_1-$locally differentially private.
	\end{lemma}
	\emph{proof.}
	Given any two possible Top-k status vector $z, z'$ and any output index $j\in\{1,\cdots, d\}$, the following conditional probability ends the proof:
	$$
	\frac{Pr[j|z]}{Pr[j|z']} 
	\le \frac{Pr[j|{z}_j=1]}{Pr[j|{z}_j'=0]}
	= \frac{p\frac{1}{k}}{(1-p)\frac{1}{d-k}} = e^{\epsilon_1},
	\text{where } p=\frac{e^{\epsilon_1}k}{d-k+e^{\epsilon_1}\cdot k}.
	$$
	\vspace{-15pt}
	
	\subsection{Analyses of Accuracy and Time Complexity}\label{subsec-time-acc}
	\noindent
	We analyze the accuracy improvement of the proposed two-stage framework by evaluating the error bound in Theorem \ref{theo-acc} which stands independently of value perturbation algorithms in the second stage.
	The amount of noise in the average vector is $O(\frac{\sqrt{\log d}}{\epsilon_2 \sqrt{md}})$ and the acceptable batch size is $|m|=\Omega(\frac{\log d}{d \epsilon_2^2})$ which does not increase linearly with $d$. Since $\epsilon_2=\epsilon^{\prime}(1-\mu)$ and $\epsilon^{\prime}=\epsilon/E$, it is evident that $\Omega(E^2 \log d/d\epsilon^2) < \Omega(E^2 d \log d/\epsilon^2)$ for $\mu<0.5$. 
	Hence, we can improve the accuracy while keeping the same private guarante. It also reminds us to allocate a small portion of privacy budget to the dimension selection.
	
	\begin{theorem}\label{theo-acc}
		For any $j\in[d]$, let $\tilde{s}=\frac{1}{m}\sum_{s^*\in \mathcal{G}}s^*$. $X=\frac{1}{m}\sum_{s^*\in \mathcal{G}} s$ denotes the mean of true sparse vectors without perturbations. With $1-\beta$ probability,
		$$\max_{j\in[1,d]}|\tilde{s}_j-X_j|=O(\frac{\sqrt{\log d/\beta}}{\epsilon_2 \sqrt{md}}).$$
	\end{theorem}
	
	Compared with the non-private setting, LDP brings extra computation costs for local devices. 
	For EXP, different utility scores and the summation can be initialized offline. Sorting a $d-$dimensional vector consumes $O(d\log d)$. Mapping all $d$ dimensions to according utility values consumes $O(d^2)$ and sampling requires $O(d)$. Thus, each local device has extra time cost  $O(d\log d+d^2+d)=O(d^2)$ for EXP.
	With a similar analysis, the extra time cost for PE is $O(d\log d+d+l)=O(d\log d)$ which is less than the time complexity $O(d^2)$ of EXP.
	Since PS avoids the perturbation for each dimension, it has a slightly less computation cost than PE with the magnitude of $O(d\log d)$. We validate this conclusion in experiments.
	
	\section{Experiments}\label{sec-experiments}
	\noindent
	In this section, we assess the performance of our proposed framework on real-world and synthetic datasets. 
	We first evaluate our selection methods without the second stage of value perturbation.
	To evaluate the improvement of reducing injected noises, we compare the learning performance with the state-of-the-art works\cite{wang2019collecting, duchi2018Minimax, shin2018Privacy} and validate theoretical conclusions in Sects.\ref{subsec-adapted} and \ref{subsec-time-acc}.
	Moreover, we implement a \textit{hyper-parameters-free} strategy that automatically initiates the budget allocation ratio $\mu$ to fit scenarios with dynamic population sizes.
	
	\subsection{Experimental Setup}
	\emph{Datasets and Benchmarks.}
	For the convenience to control data sparsity, the synthetic data is generated with the existing procedure\cite{wangni2018gradient} with two parameters $C_1=\{0.01,0.1\}, C_2=\{0.6,0.9\}$ and dimensions \{100(syn-L), 300(syn-H)\} over \{60,000, 100,000\} records.
	The over real-world benchmark datasets includes BANK, ADULT, KDD99 which have \{32, 123, 114\} dimensions over \{45,211, 48,842, 70,000\} records. We follow a typical pre-process procedure in machine learning with one-hot-encoding every categorical attribute. 
	We test on $l2$-regularized Logistic Regression and Support Vector Machine. 
	
	\emph{Choices of Parameters.} 
	Since we observe that models on the above datasets can converge within 100 rounds of iterations, we set the batch size for one global model's iteration as $m=0.01\cdot N$. We report the average accuracy or misclassification rates of 10 times 5-folds cross-validations for one epoch unless otherwise stated. 
	The discounting factor $\eta$ and learning rate $\alpha$ are same in each case for a fair comparison. We set $k=0.1d$, $\mu=0.1$, $\lambda=0.0001$ by default.
	
	\emph{Comparisons with Competitors.}
	The proposed \textit{FedSel} framework is prefixed with selection mechanisms EXP/PE/PS.
	We compare it with non-private baselines (NP) of three different transmitting methods (-/RS/K): full gradient, random sampled dimension, Top-k($k=1$) selection.
	We also compare with the flat solution\cite{wang2019collecting, duchi2018Minimax} and the compressed solution\cite{shin2018Privacy} with random sampling and random projection respectively before perturbing the value. 
	Due to limited space and variant baselines, we mainly demonstrate comparisons with the optimal competitor PM and show comparisons with other competitors in Fig.\ref{fig-RP} and Fig.\ref{fig-HM-Duchi}.
	Abbreviations of different variants are listed in Table \ref{tab-frameworks}.
	
	\begin{table}[t]
		\centering
		\caption{Frameworks and Variants for Comparisons.}\label{tab-frameworks}
		\begin{tabular}{|c|c|c|c|c|}
			\hline
			solution & abbreviation &sparsification & perturbation & budget \\
			\hline
			non-private & NP & full/random/topk & - & $\infty$\\
			\hline
			flat\cite{wang2019collecting, duchi2018Minimax} & PM/HM/Duchi & random sampling & $\epsilon^{\prime}$ & $\epsilon^{\prime}$\\
			\hline
			compressed\cite{shin2018Privacy} & -RP & random projection & $\epsilon^{\prime}$ &$\epsilon^{\prime}$\\
			\hline
			two-stage & EXP/PE/PS- &$\epsilon_1=\mu \cdot \epsilon^{\prime}$ & $\epsilon_2=\epsilon^{\prime}-\epsilon_1$ & $\epsilon^{\prime}$\\
			\hline
		\end{tabular}
	\end{table}	

	\subsection{Evaluation of the Dimension Selection}
\begin{figure}[t]
		\subfigure[]{\label{fig-converge-L}
			\centering
			\includegraphics[width=0.3\textwidth,trim=0 0 0 0,clip]{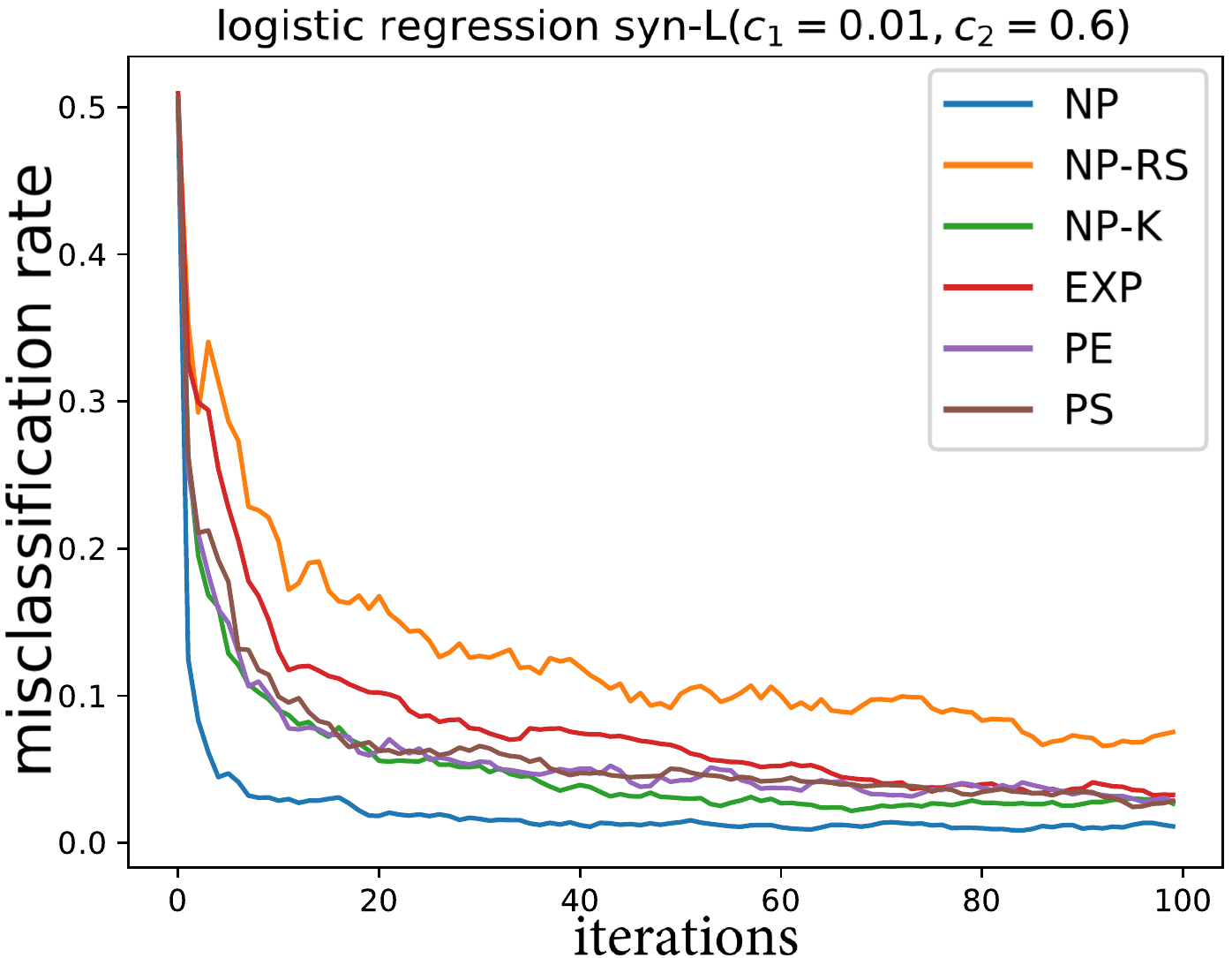}
		}
		\subfigure[]{\label{fig-converge-H}
			\centering
			\includegraphics[width=0.3\textwidth,trim=0 0 0 0,clip]{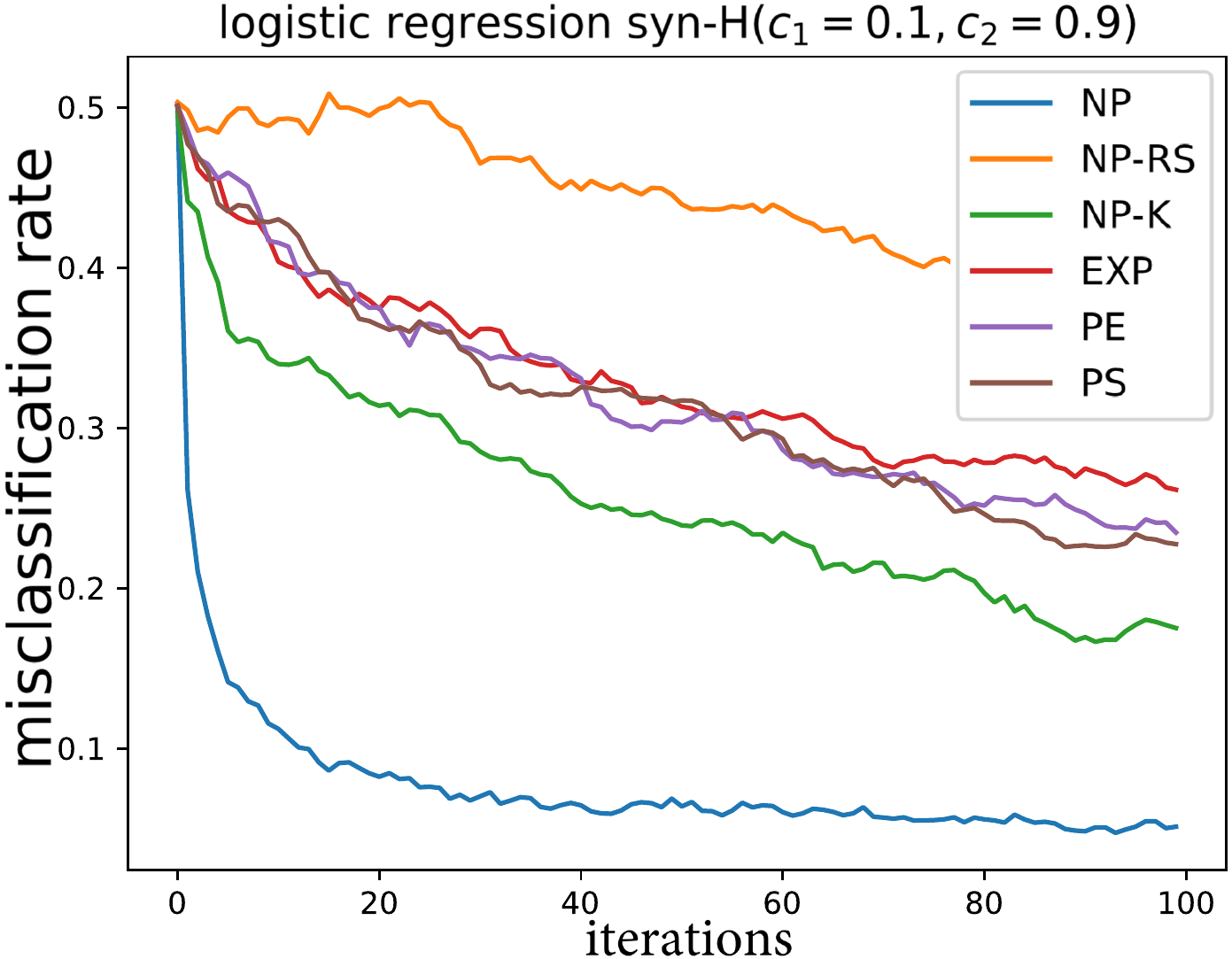}
		}
		\subfigure[]{\label{fig-acc-L}
			\centering
			\includegraphics[width=0.3\textwidth,trim=0 0 0 0,clip]{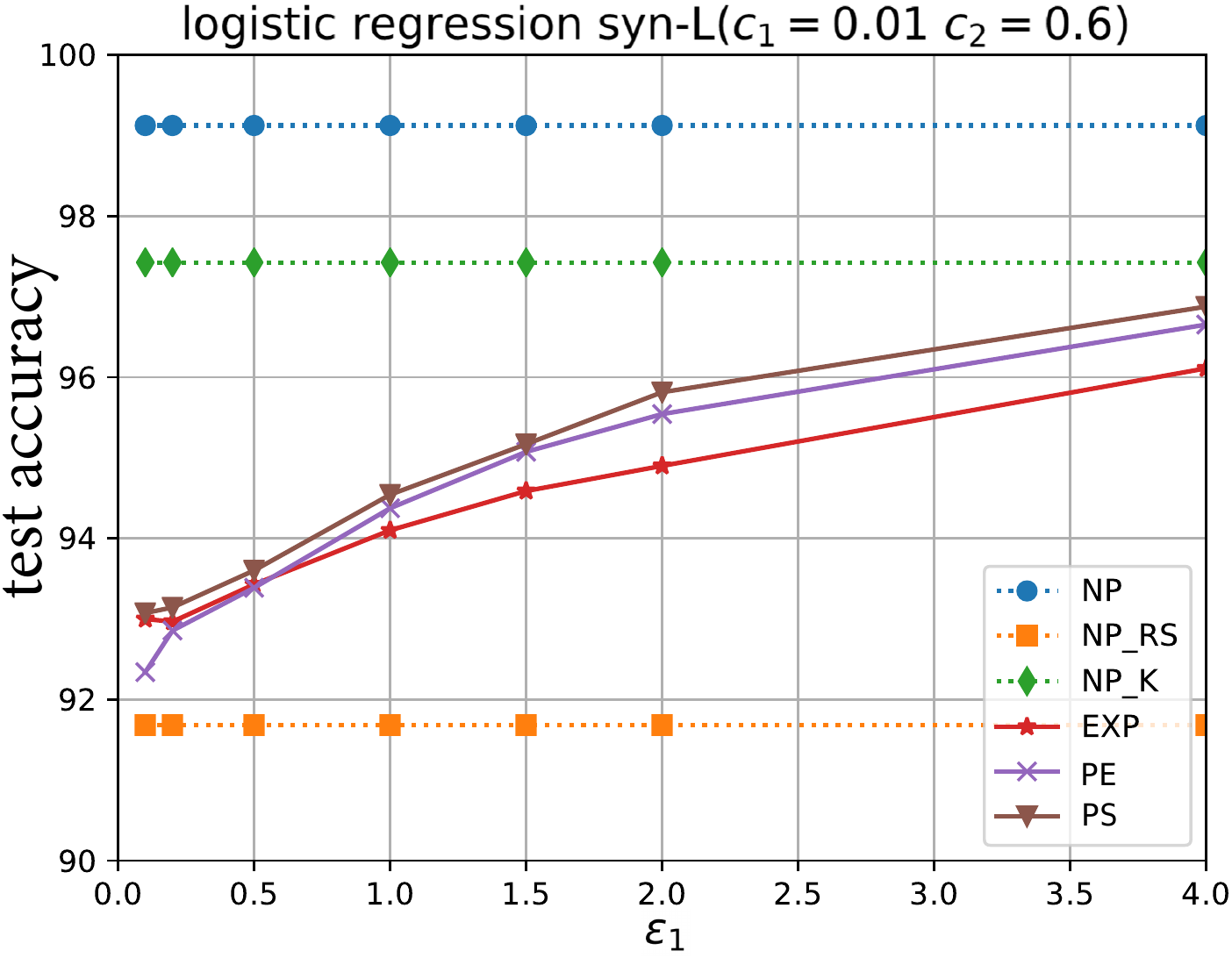}
		}
		\subfigure[]{\label{fig-acc-H}
			\centering
			\includegraphics[width=0.3\textwidth,trim=0 0 0 0,clip]{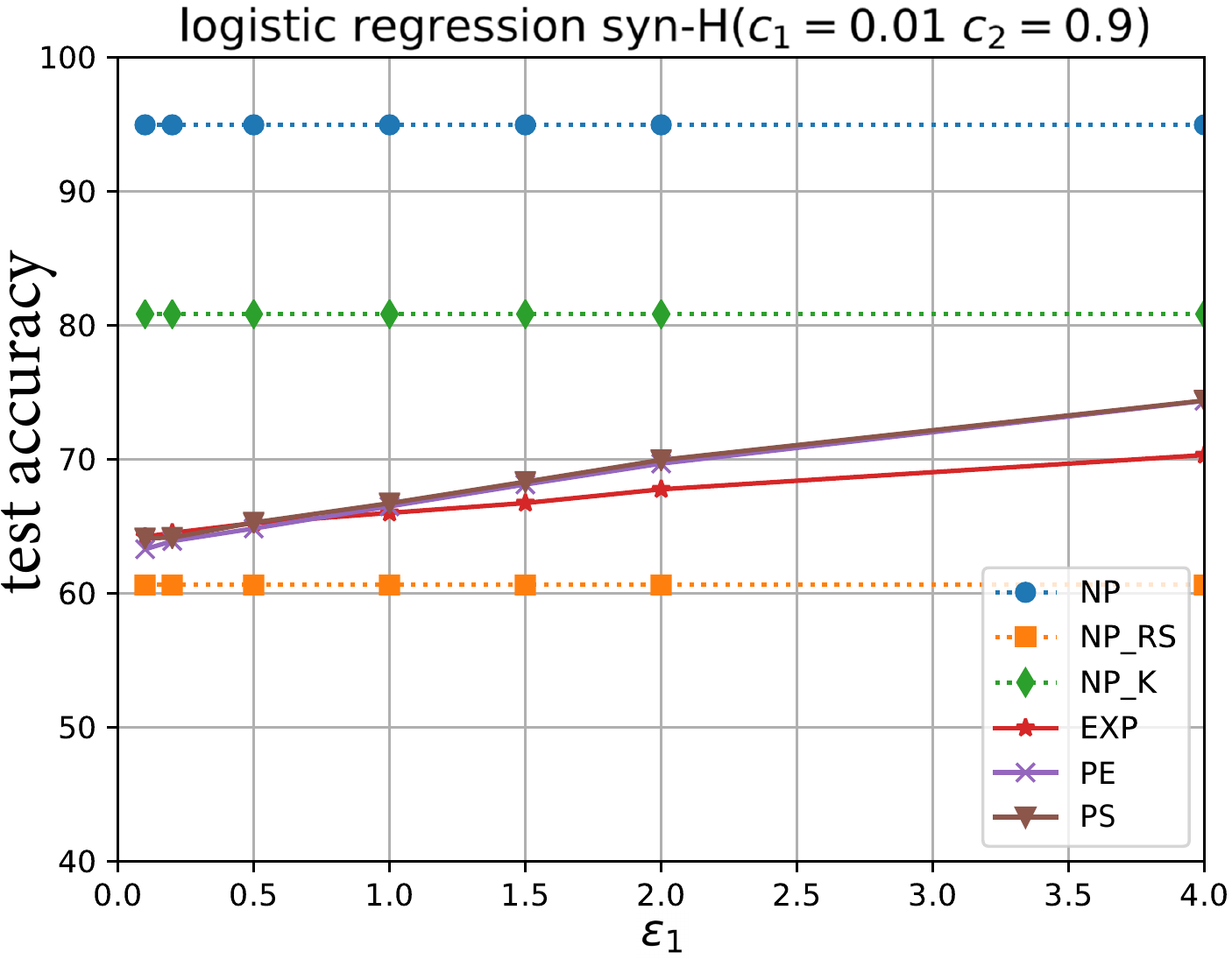}
		}
		\subfigure[]{\label{fig-time}
			\centering
			\includegraphics[width=0.3\textwidth,trim=0 0 0 0,clip]{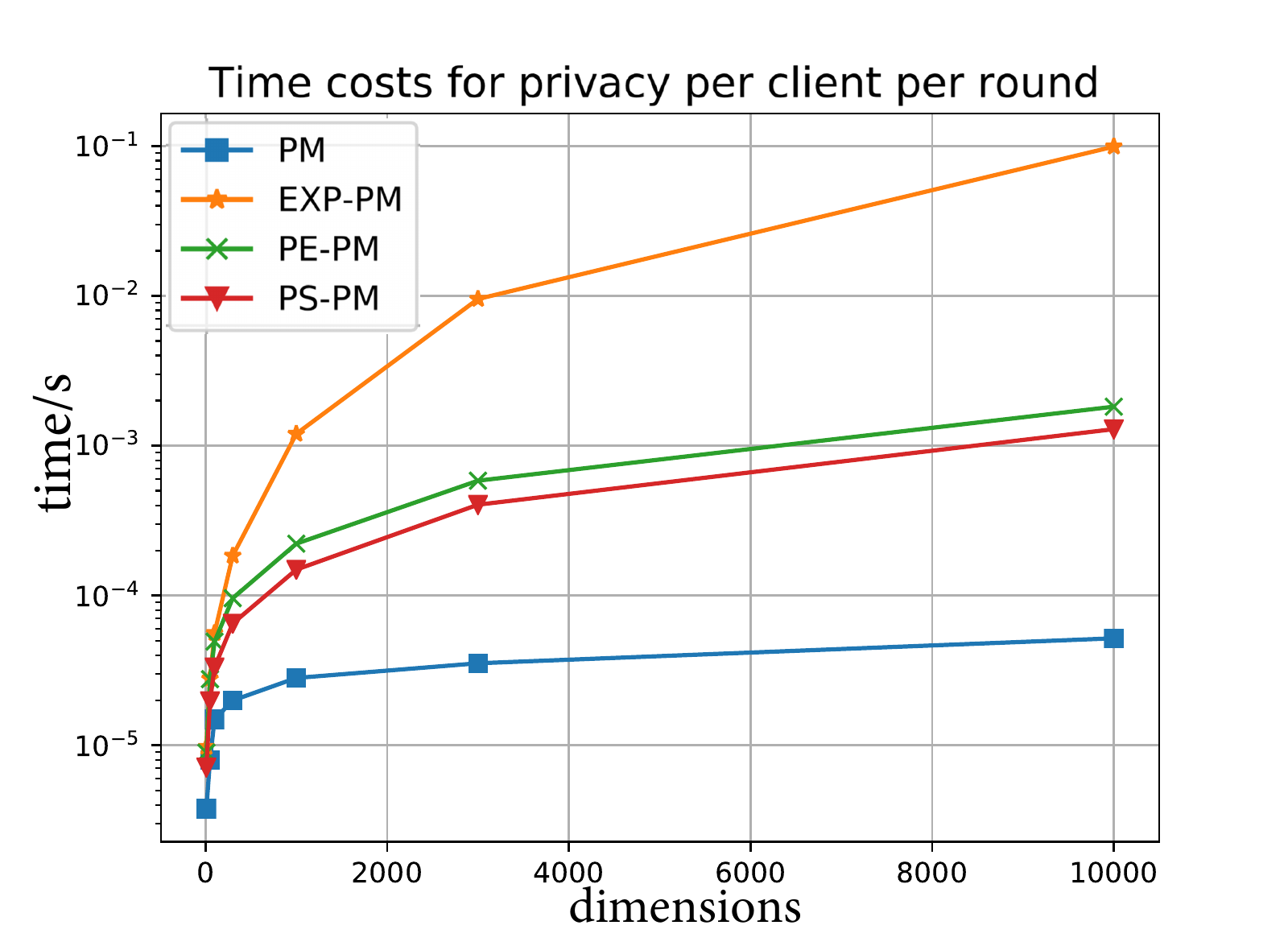}
		}
		\subfigure[]{\label{fig-bank}
			\centering
			\includegraphics[width=0.3\textwidth,trim=0 0 0 0,clip]{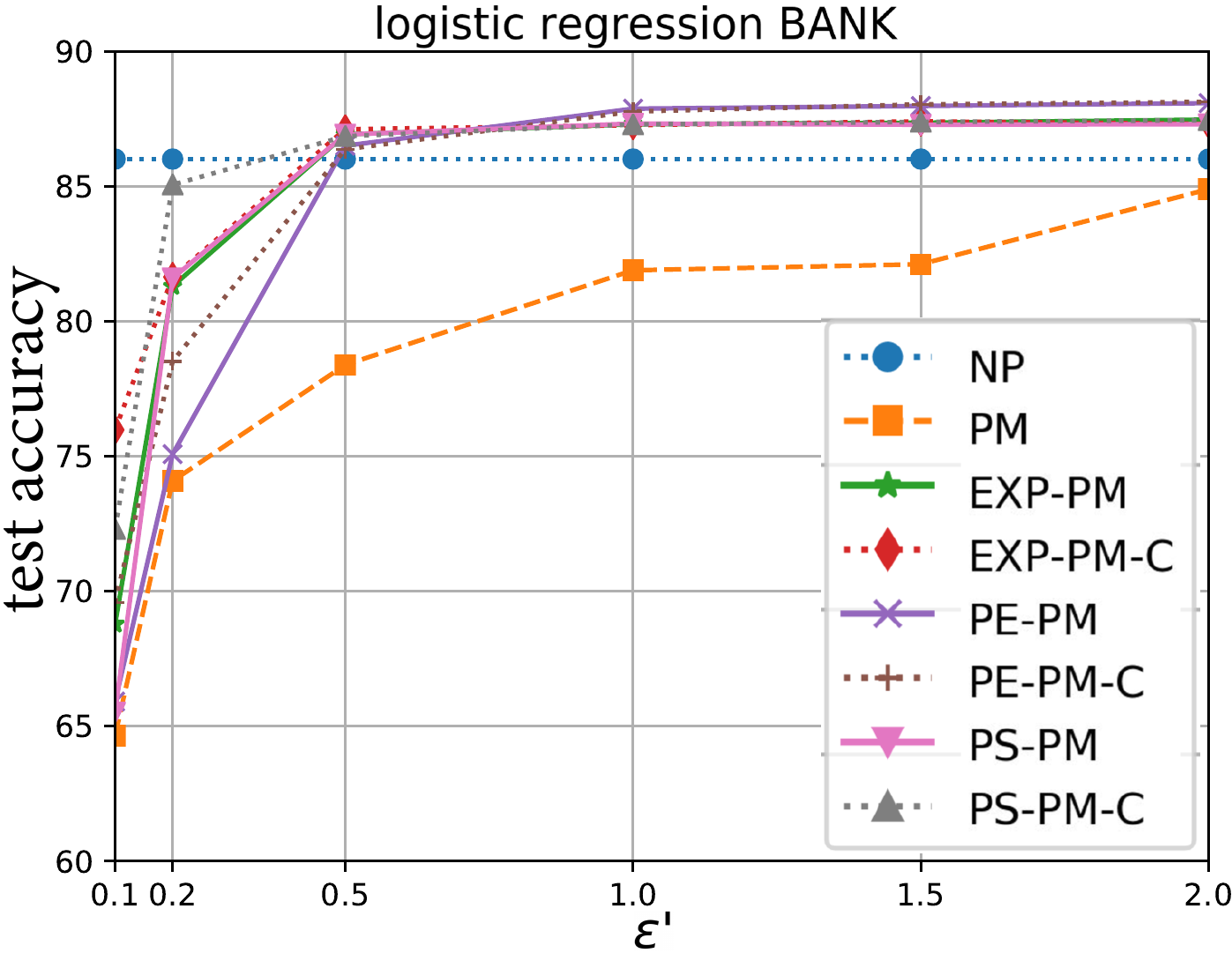}
		}
		\subfigure[]{\label{fig-kdd}
			\centering
			\includegraphics[width=0.31\textwidth,trim=10 5 28 8,clip]{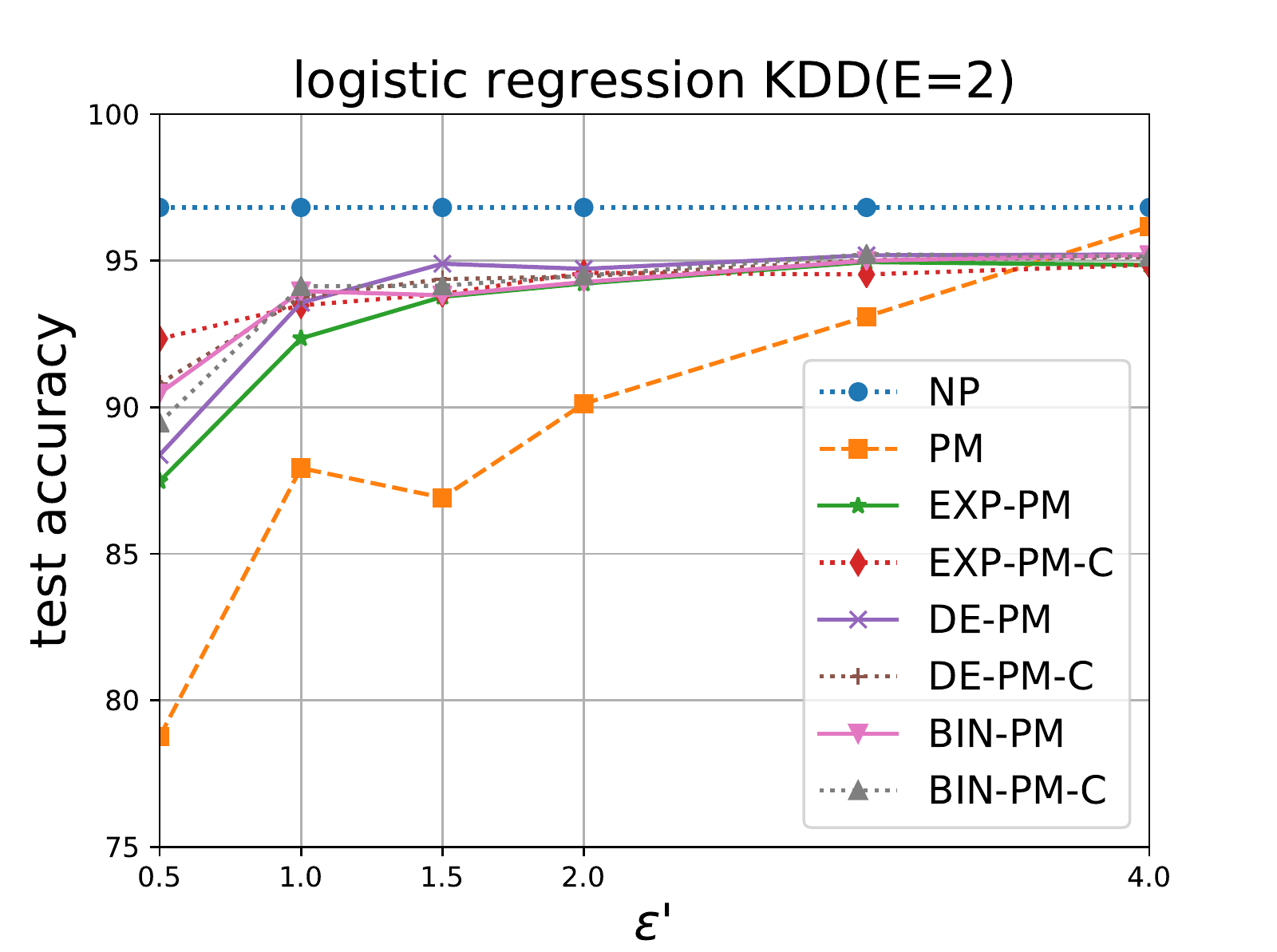}
		}
		\subfigure[]{\label{fig-HM-Duchi}
			\centering
			\includegraphics[width=0.31\textwidth,trim=0 0 0 0,clip]{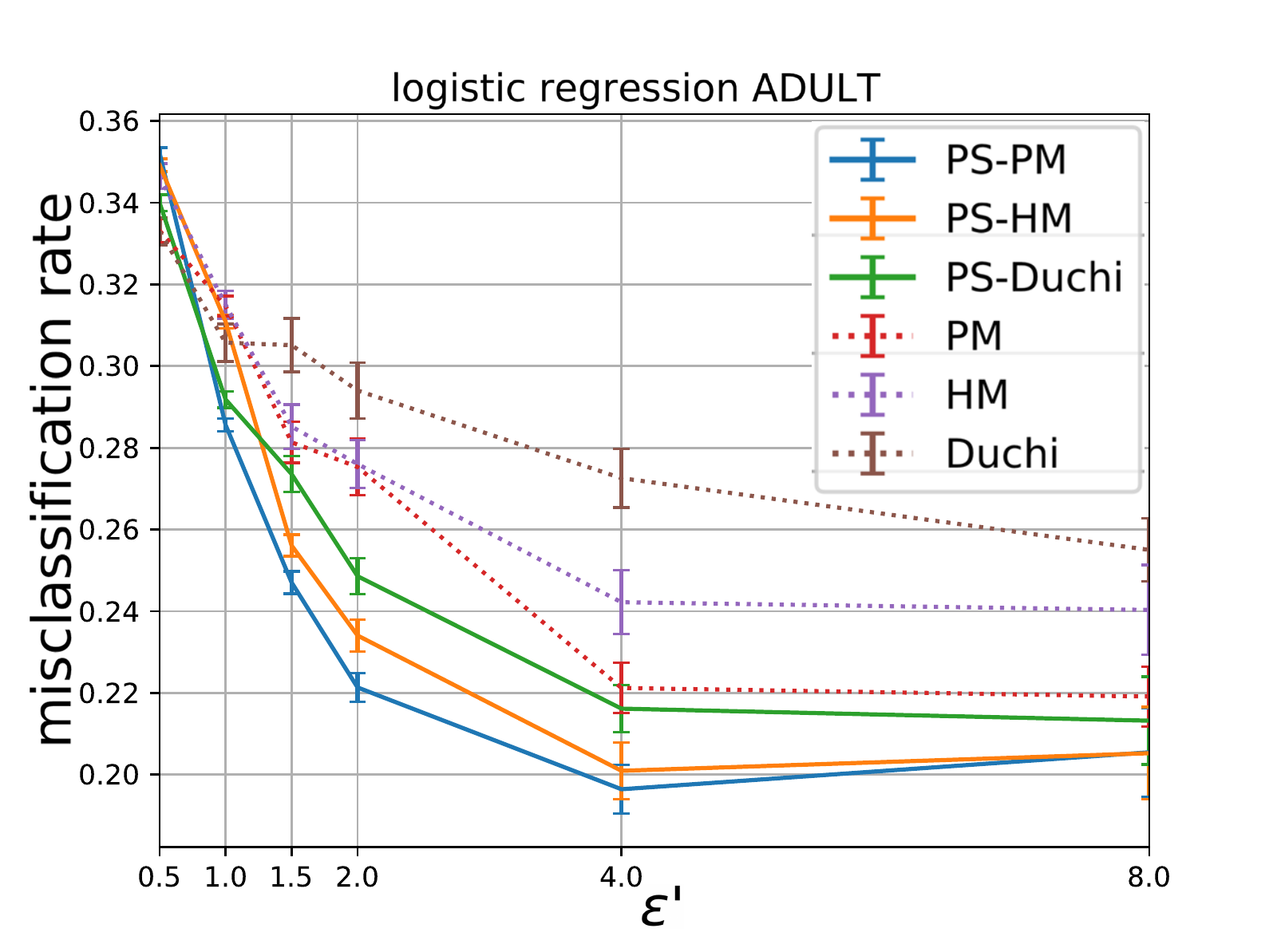}
		}
		\subfigure[]{\label{fig-RP}
			\centering
			\includegraphics[width=0.31\textwidth,trim=2 4 38 2,clip]{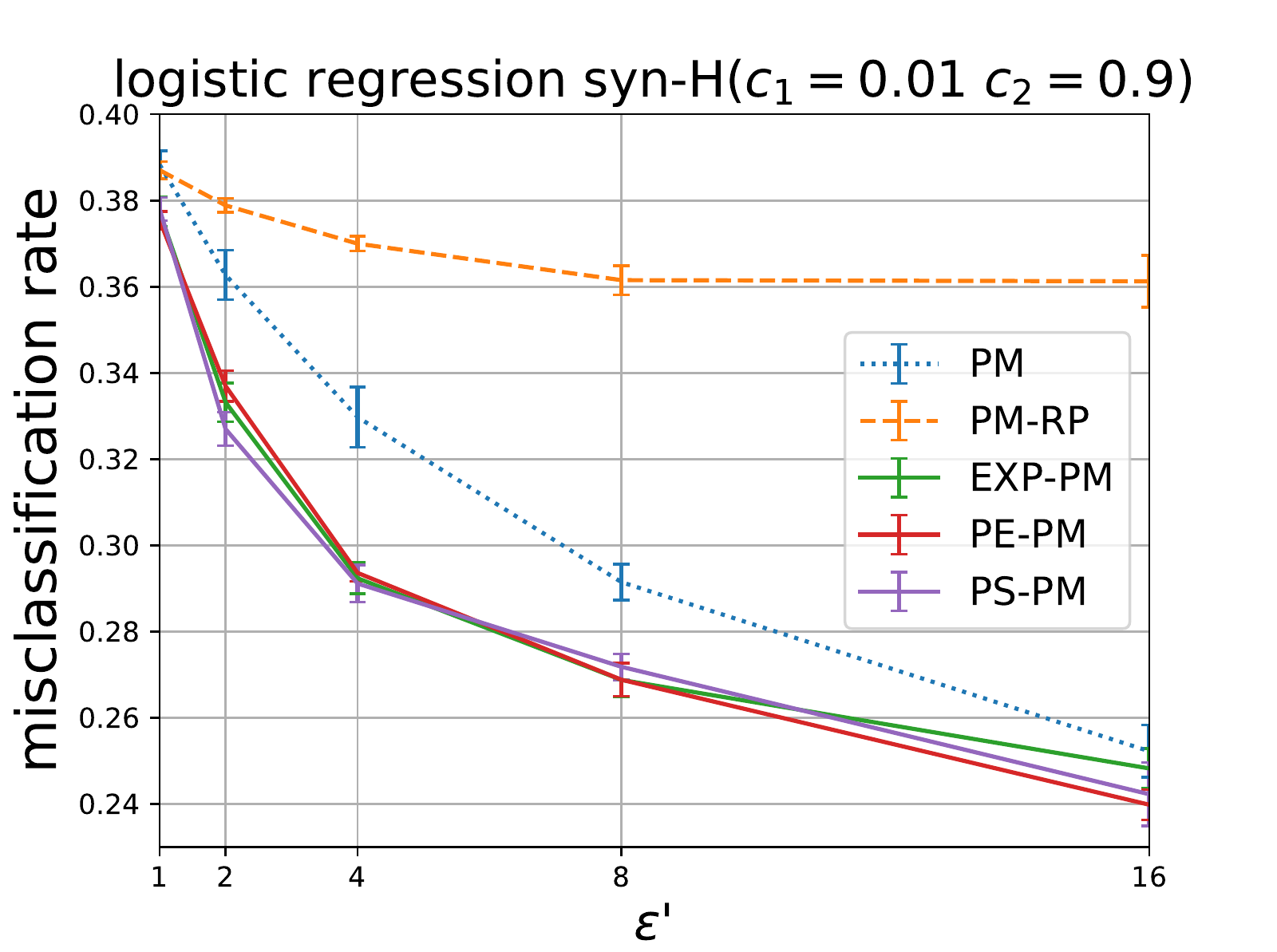}
		}
		\caption{Improvements of the two-stage framework and dimension selection.}
	\end{figure}

	\emph{Convergence and Accuracy.} We compare EXP/PE/PS with non-private baselines of NP, NP-RS and NP-K by visualizing the misclassification rate and accuracy of test set in Fig.\ref{fig-converge-L} to Fig.\ref{fig-acc-H}. 
	Note that this comparison only focuses on selection without the second stage of value perturbation.
	For NP-RS, we enlarge the value randomly sampled by each user for an unbiased estimation. This follows the same principle in the flat or compressed solution.
	
	The advance of NP-K compared with NP-RS shows our essential motivation that Top-k is a more effective and accurate way to reduce the transmitted dimension. On 100-dimensional dataset, NP-K even approaches the full-gradient-uploading baseline NP in Fig.\ref{fig-converge-L}.
	Besides, EXP/PE/PS converge more stable and faster than NP-RS in Fig.\ref{fig-converge-L} and Fig.\ref{fig-converge-H} with $\epsilon_1=4$.
	With a larger privacy budget, there is a trend for EXP/PE/PS to approach the same accuracy performance as the NP-K($k=1$) in Fig.\ref{fig-acc-L} and Fig.\ref{fig-acc-H}. Moreover, even a small budget in dimension selection helps to increase the learning accuracy. 
	We can also observe that PE and PS methods which intuitively intend to select from the Top-k list have a better performance than EXP. Thus, our extension from Top-1 to Top-k is necessary.
	
	\emph{Validation of Complexity Analysis.}
	Here we analyze the time consumption for each client per transmission in Fig.\ref{fig-time}. The time is counted by iterating over synthetic datasets with variant dimensions from 10 to 10,000. We observe that the selection stage indeed incurs extra computation cost. Consistent with previous analysis in Sect.\ref{subsec-time-acc}, PS has the lowest computation cost.
	
	\subsection{Effectiveness of the Two-stage Framework}
	\emph{Comparison with Existing Solutions.}
	We compare the learning performance for EXP/PE/PS-PM with the flat solution PM/HM/Duchi\cite{wang2019collecting, duchi2018Minimax} and the compressed solution\cite{shin2018Privacy}.
	Remark that we set control groups with postfix "-C" in Fig.\ref{fig-bank} and Fig.\ref{fig-kdd} by allocating $\epsilon_1$ for dimension selection and $\epsilon^{\prime}$ for value perturbation. 
	To further elucidate the trade-off between what we gain and what we lose, we qualify the benefit of private selection with the gap between the accuracy as the following, which is shown as Table \ref{tab-gain-loss}.
	It is evident that what we gain is much larger than what we lose. This is because when we have enough privacy budget for value perturbation, increasing budget for value perturbation is not comparable to allocating surplus budget to privacy selection.
	\begin{align*}
	\indent \text{gain} &= \text{acc}(\text{EXP/PE/PS-PM-C})-\text{acc}(\text{PM}),\\
	~~\text{loss} &= \text{acc}(\text{EXP/PE/PS-PM-C})-\text{acc}(\text{EXP/PE/PS-PM}).
	\end{align*}
	
	\begin{table}[t]
		\caption{Gains and Losses on Accuracy for Private Selection with $\epsilon=2$  (\%).}
		\begin{center}\label{tab-gain-loss}
			\begin{tabular}{|c|c|c|c|c|c|c|c|}
				\hline
				dataset & model &EXP-gain & EXP-loss & PE-gain & PE-loss & PS-gain & PS-loss\\
				\hline
				syn-L-0.01-0.9 & logistic & \textbf{8.6074} & 0.3517 & \textbf{5.410} & 1.192 & \textbf{5.975} &0.4970 \\
				\hline
				syn-L-0.01-0.9 & SVM & \textbf{7.1950} & 2.1593 & \textbf{3.7704} & 0.8533 & \textbf{5.065} &2.0816 \\
				\hline
				BANK & logistic & \textbf{2.4197} & -0.157 & \textbf{3.2338} & 0.0464 & \textbf{2.5525} &0.1463 \\
				\hline
				BANK & SVM & \textbf{4.3823} & 0.4436 & \textbf{3.4369} & 0.2530 & \textbf{4.0244} &0.0164 \\
				\hline
				KDD & logistic & \textbf{2.0471} & 0.5091 & \textbf{2.5148} & 0.2322 & \textbf{2.0171} &0.3428 \\
				\hline
				KDD & SVM & \textbf{1.85629} & -0.1625 & \textbf{2.2168} & 0.2288 & \textbf{1.8291} &0.4465 \\
				\hline
				ADULT & logistic & \textbf{5.5745} & 0.2935 & \textbf{5.6445} & 1.3096 & \textbf{6.0535} &0.8091 \\
				\hline
				ADULT & SVM & \textbf{5.5361} & 0.1949 & \textbf{5.6057} & 0.9550 & \textbf{5.1442} &0.3852 \\
				\hline
			\end{tabular}
		\end{center}
	\end{table}
	
	From Fig.\ref{fig-bank} and Fig.\ref{fig-kdd}, we observe that proposed two-stage solutions have higher test accuracy than the optimal private baseline PM on both models and all datasets. Given enough privacy budget on relatively low-dimensional datasets, proposed solutions even outperform the non-private baseline in Fig.\ref{fig-bank}. 
	The key to this success is the inherent randomness in SGD. The slightly introduced stochasticity for privacy-preserving prevents the overfitting problem.
	In Fig.\ref{fig-kdd} with results of two epochs, our adapted local accumulation with momentum helps to reduce the impact of noisy gradients compared with the private baseline PM, especially when $\epsilon^{\prime}$ is small.
	
	From Fig.\ref{fig-HM-Duchi}, we show a comparison with flat solutions of other perturbation methods which have the same optimal error bound as PM. 
	Note that the value perturbation algorithms for each pair of comparison in Fig.\ref{fig-HM-Duchi} is the same and only differ in the selection stage.
	It is evident that our framework has a lower misclassification rate and standard deviation over 50 times tests. 
	Therefore, we can conclude that this improvement is independent of value perturbation methods. We omit the comparison of EXP and PE for the same conclusion.
	
	In Fig.\ref{fig-RP}, we compare with the compressed solution\cite{shin2018Privacy} with the same comparison ratio 0.1. 
	The originally apply random projection in gradients of the Matrix Factorization and use another value perturbation method\cite{nguyen2016collecting}. Since the dimension reduction idea is independent of value perturbation, for fairness comparison, we adopt the random projection idea and use the same method PM\cite{wang2019collecting} to perturb value when implementing the competitor PM-RP. Our result shows that, even the error bound is reduced by random projection, the recovery error ruins the accuracy while our dimension selection still works.

	\begin{figure}[htb]
		\subfigure[]{\label{fig-adapted}
			\includegraphics[width=0.31\textwidth,trim=0 0 0 0 ,clip]{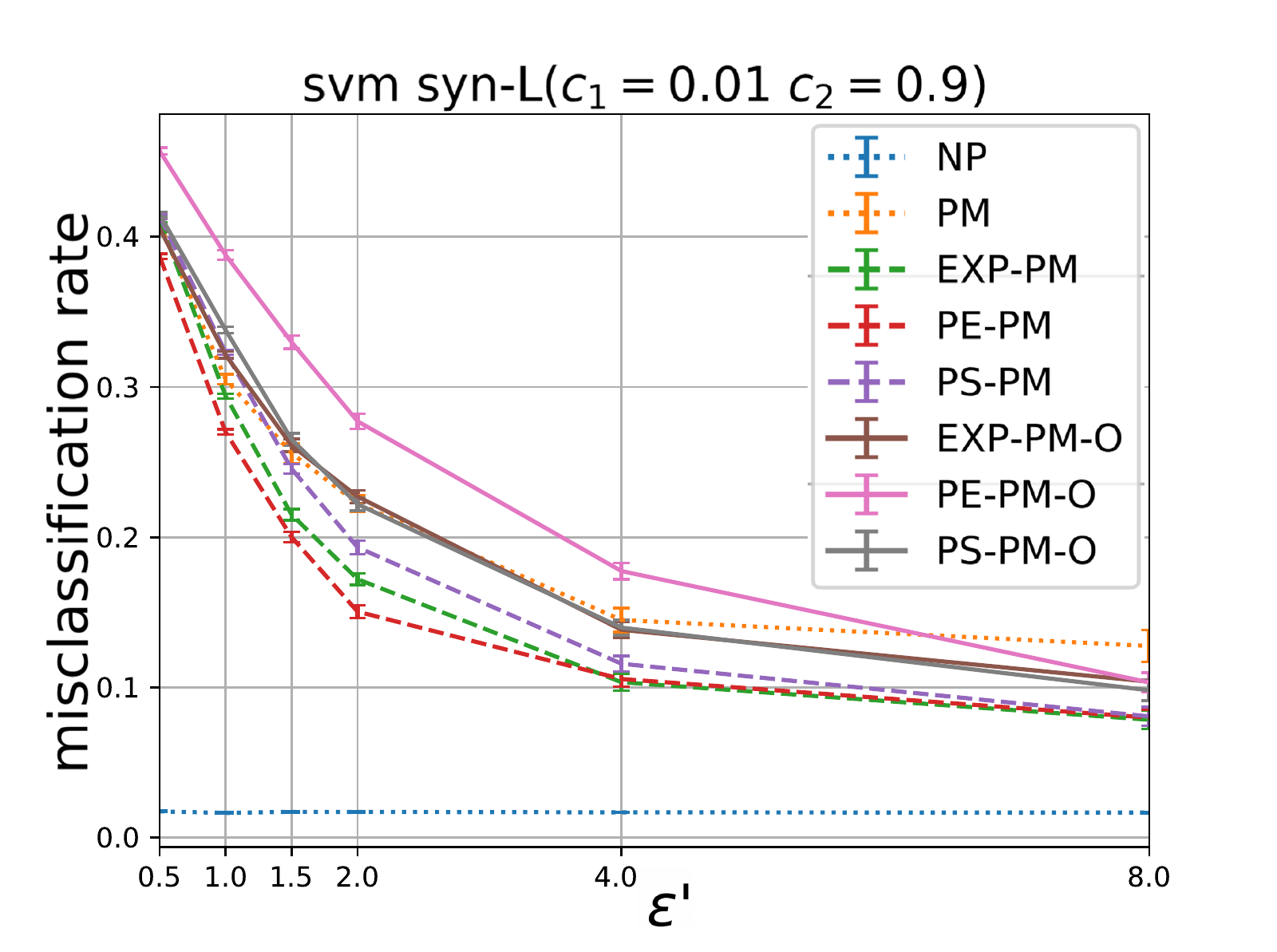}
		}
		\subfigure[]{\label{fig-mu}
			\includegraphics[width=0.31\textwidth,trim=0 0 0 0 ,clip]{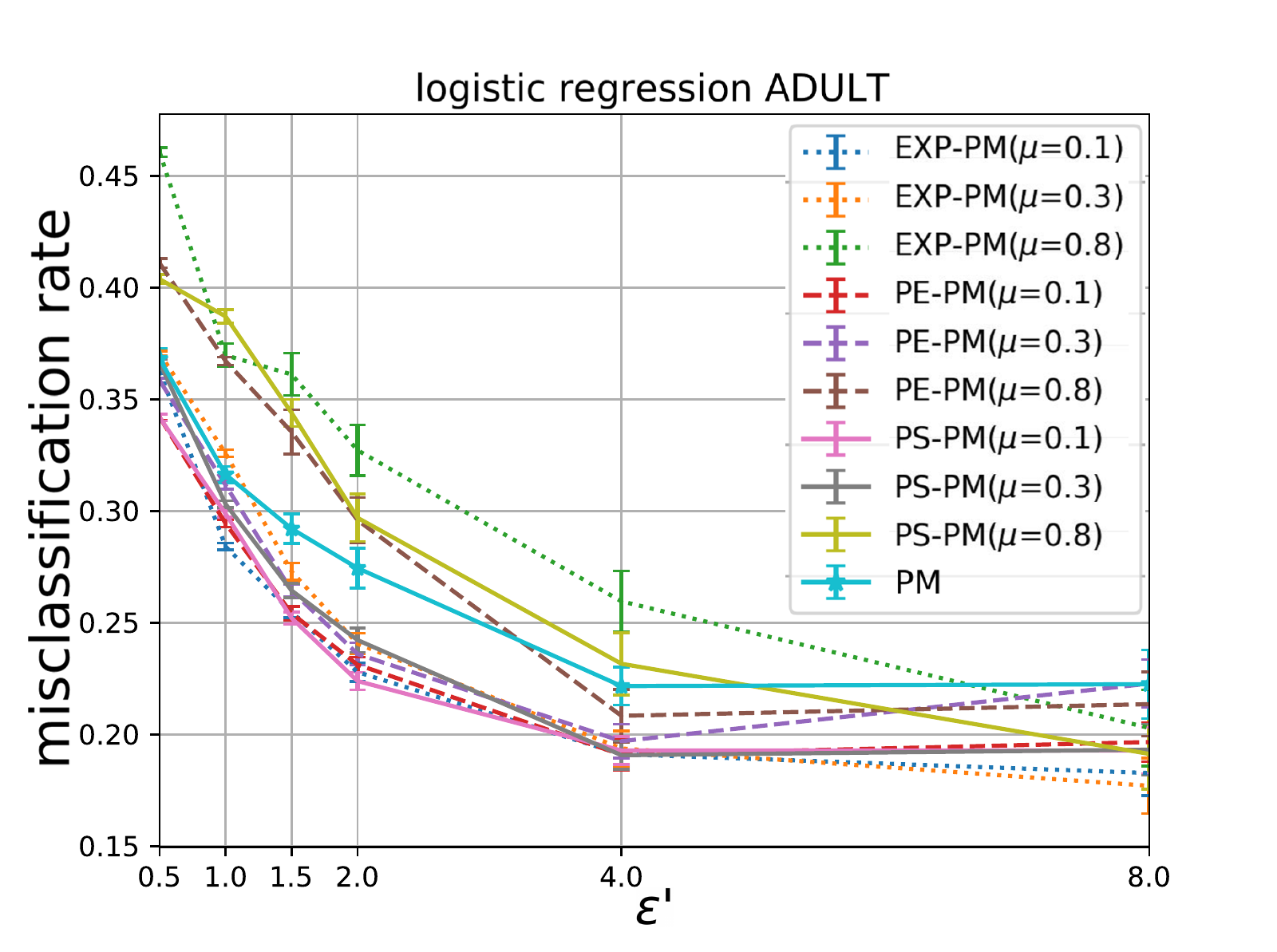}
		}
		\subfigure[]{\label{fig-hpf}
			\includegraphics[width=0.31\textwidth,trim=2 4 35 2,clip]{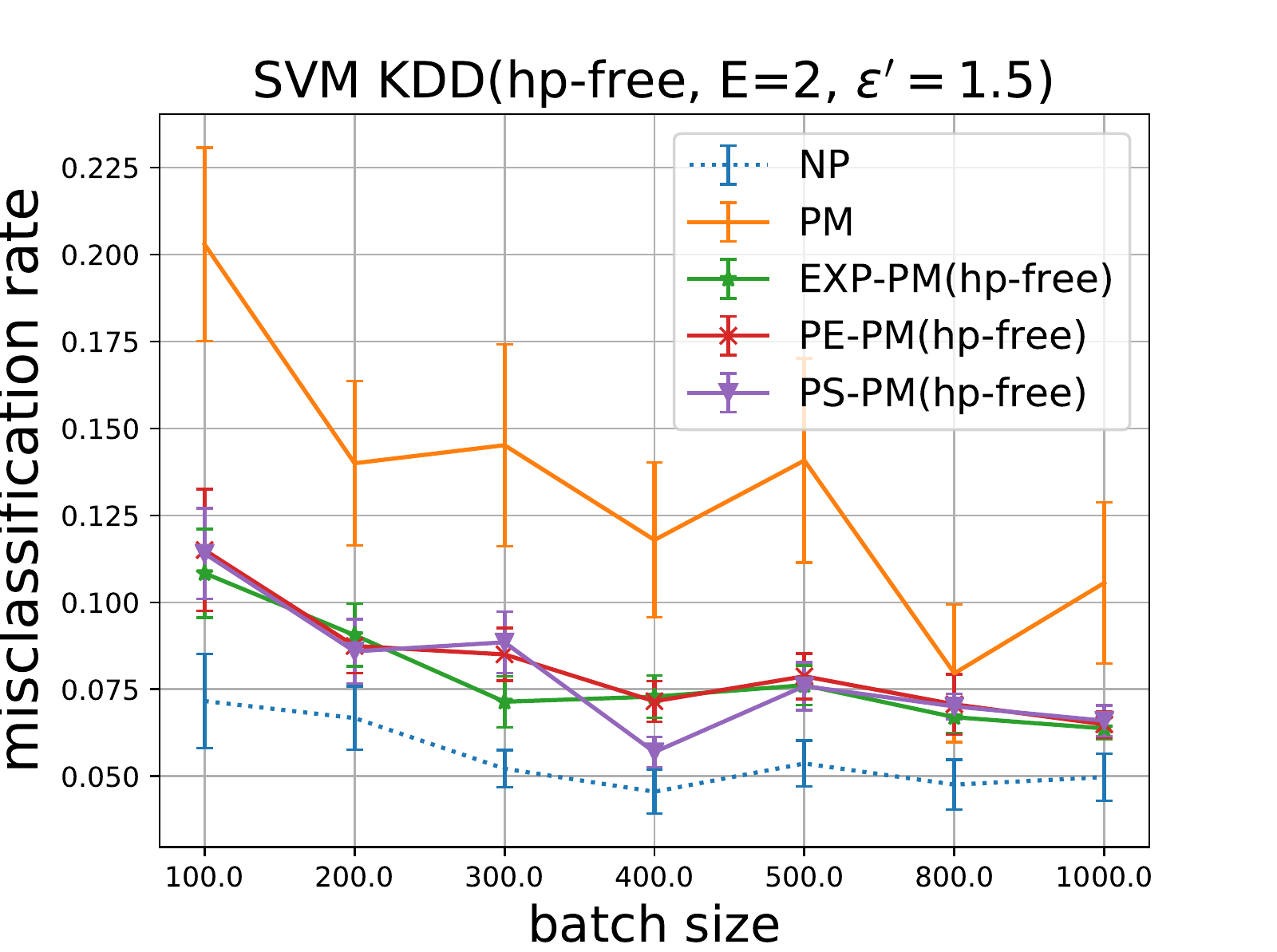}
		}
	\caption{Improvement of the adapted accumulation and impacts of $\mu$}
	
	\end{figure}
	
	\emph{Effectiveness of the Adapted Accumulation.}
	We validate the improvement of stability consistent with our analysis in Sect.\ref{subsec-adapted} in Fig.\ref{fig-adapted}. 
	We set the variant of accumulating $\alpha g$ as the competitor (EXP/PE/PS-PM-O). The result shows accumulating $\alpha g$ directly will not improve the learning performance because what we gain by selection is offset by the larger turbulence.
	Thus, our adaptation is necessary for better compatibility with the private context.
	
	\emph{Impacts of $\mu$.}
	As $\mu \in[0,1]$ controls the privacy budget allocation in our two-stage framework, we evaluate $\mu$ in Fig.\ref{fig-mu} with ADULT dataset while the same trends are found in other datasets.
	From Fig.\ref{fig-mu}, we observe that our framework with a small $\mu$ works no worse than the flat competitor, even when the total budget is small. In addition, $\mu =0.1$ gives an optimal learning accuracy, and $\mu=0.8$ leads to a worse performance with a higher misclassification rate and a significant standard deviation as expected.
	Thus, $\mu$ is an essential parameter that controls the trade-off between what we gain and what we lose.
	
	The divergence of a large $\mu$ reminds us that a safe maximum threshold $\theta$ is required to guarantee this trade-off always benefits the model's accuracy.
	It is much easier to tune $\theta$ than $\mu$ as $\theta$ can be tested on synthetic datasets independently of total privacy budget and batch size.
	At the beginning of training, given the privacy budget per epoch $\epsilon'$, the model's dimension $d$, the available batch size $m$, our principle is to first allocate at least $\epsilon_2=\Omega(\sqrt{d \log d/m})$ to the second stage. Then extra privacy budget can be allocated for dimension selection to improve the accuracy as a bonus.
	
	In Fig.\ref{fig-hpf}, we set a safe threshold as $\theta=0.2$ empirically and validate the effectiveness of the proposed hyper-parameters-free strategy. 
	Different batch sizes shown in the x-axis simulate the dynamic amount of participants when initiating a distributed learning task in practice. 
	For the fairness to compare the test set accuracy with different batch sizes, we stop the learning process within the same number of iterations.
	We observe that the proposed solution with three dimension selection methods under this strategy significantly improve the model's accuracy. 
	Besides, the proposed \textit{hyper-parameters-free} strategy works steadily for dynamic batch sizes as the deviations among all 50 times tests are smaller than the private baseline. 
	
	\section{Related Works}\label{sec-related works}
	\noindent
	How we select the dimension and accumulate gradients are based on the well-studied gradient sparsification in the non-private setting. 
	Strom et al.\cite{strom2015scalable} propose only to upload dimensions with absolute values larger than a threshold. Instead of a fixed threshold, Aji et al.\cite{aji2017sparse} introduce Gradient Dropping (GD) which sorts absolute values first and dropping a fixed portion of gradients. Wang et al.\cite{wangni2018gradient} drop gradients by trading off between the sparsity and variance. Alistarh et al.\cite{alistarh2018convergence} show the theoretical convergence for Top-k selection.
	However, even if the gradient update is compressed, there still exist privacy risks because it is calculated directly with local data.
	
	If local gradients are transmitted in clear, the untrusted server threatens clients' privacy.
	Nasr et al.\cite{nasr2019comprehensive} present the membership inference by only observing uploads or controlling the view of each participant. Wang et al.\cite{wang2019beyond} propose a reconstruction attack in which the server can recover a specific user's local data with Generative adversarial nets (GANs). Secure attack\cite{fang2019local} in FL is also an important topic but we focus on private issues in this paper.
	
	Cryptography technologies face a bottleneck of heavy communication and computation costs. Bonawits et al.\cite{bonawitz2017Practical} present an implementation of Secure Aggregation, which entails four rounds interacts per iteration and several costs grow quadratically with the number of users. As for differential privacy(DP)\cite{dwork2014the(book)} in distributed SGD, Shokri et al.\cite{shokri2015privacy} propose a asynchronous cooperative learning with privately selective SGD by sparse vector technique. It may lose accuracy as it drops delayed gradients instead of accumulating as our works. Agarwal et al.\cite{agarwal2018cpsgd} combine gradient quantization and differential private mechanisms in synchronous setting, but it requires a higher communication cost for $d$-dimensional vector. It should be noticed that the privacy definition in the above works differentiate from LDP as it provides the plausible deniability for only single gradient value while LDP guarantees the whole gradient vector to be indistinguishable.
	
	Many LDP techniques are proposed for categorical or numeric values. Randomized response (RR) method\cite{warner1965randomized} is the classic method to perturb binary variables. Kairouz et al.\cite{kairouz2014extremal} introduce a family of extremal privatization mechanisms $k$-RR to categorical attributes. With LDP mechanisms for mean estimation, Duchi et al.\cite{duchi2018Minimax} suggest that LDP leads to an effective degradation in batch size. Wang et al.\cite{wang2019collecting} show that compared with Duchi et al.'s work, their mean estimation mechanisms with lower worst-case variance lead to a lower misclassification rate when applied in SGD. Considering the required batch size is linearly dependent on the dimension, Bhowmick et al.\cite{bhowmick2018protection} design LDP mechanisms for reconstruction attack with a large magnitude of privacy budget to get rid of the utility limitation of a normal locally differentially private learning.
	
	\section{Conclusions}\label{sec-conclusions}
	\noindent
	This paper proposes a two-stage LDP privatization framework \textit{FedSel} for federated SGD. 
	The key idea takes the first attempt to mitigate the dimension problem in injected noises by delaying unimportant gradients.
	We further stabilize the global iteration by modifying the accumulation with a smaller variance on the noisy update.
	The improvement of proposed methods is theoretically analyzed and validated in experiments. The framework with \textit{hyper-parameters-free} also outperforms baselines over variant batch sizes.
	In future work, we plan to formalize the optimal trade-off for utility and accuracy and extend \textit{FedSel} to a more general case.\\
	
	\noindent
	\textbf{Acknowledgements.} This work is supported by the National Key Research and Development Program of China (No. 2018YFB1004401), National Natural Science Foundation of China (No. 61532021, 61772537, 61772536, 61702522), JSPS KAKENHI Grant No. 17H06099, 18H04093, 19K20269, and Microsoft Research Asia (CORE16).

	
	\newpage
	\begin{subappendices}
		\section{Analysis of Accuracy}\label{appendix-accuracy}
		\emph{proof. for Theorem \ref{theo-acc}}: 
		For any $j \in [1, d]$, the estimated mean of perturbed sparse vectors is $\tilde{s}=\frac{1}{m}\sum_{s^*\in \mathcal{G}}s^*$. The mean of the sparse vectors without value perturbation is denoted as $X=\frac{1}{m}\sum_{s^*\in \mathcal{G}} s$.
		For any $i \in [1,m]$, perturbation method is unbiased, $s^*_{i,j}-s_{i,j}$ has zero mean.
		Given privacy budget $\epsilon_2$ for value perturbation with the asymptotically optimal error bound, we have the upper bound for $|s^*_{i,j}-s_{i,j}|$ as $O(1/\epsilon_2)$.
		Then we have:
		\begin{align}
		Pr&[|\tilde{s}_j-X_j|\ge \lambda] \notag\\
		=& Pr[|\sum_i^m \{s^*_{i,j}-s_{i,j}\}|\ge m\lambda] \notag\\
		\le& 2 \cdot \exp {(-\frac{(m\lambda)^2}{2\sum_{i}^m Var[s^*_{i,j}]+\frac{2}{3} m \lambda \cdot O(1/\epsilon_2)})} \notag \tag{1}\label{tag1}
		\end{align}
		Let $p_{i,j}$ denote the probabilities for dimension $j$ of user $u_i$ being selected by a selection mechanism. $\zeta$ presents the output value of the value perturbation algorithm, and we have $\mathbb{E}[\zeta^2]=O(1/\epsilon_2^2)$.
		Thus, the expectation of the variance is shown as follows:
		\begin{align}
		Var[s_{i,j}^*]
		=&\mathbb{E}[(s_{i,j}^*)^2]-\mathbb{E}[s_{i,j}^*]^2 \notag\\
		=&p_{i,j} \mathbb{E}[\zeta^2]-(s_{i,j})^2 \tag{2}\label{tag2}.
		\end{align}
		
		The probability of $p_{i,j}$ depends on whether $j$ is in the Top-k list of $u_i$'s local vector. For selection mechanisms EXP, PE, PS, we have the upper bound of $p_{i,j}$ dominated by $O(1/d)$.
		Apply (\ref{tag2}) to (\ref{tag1}), we have:
		\begin{align*}
		Pr&[|\tilde{s}_j-X_j|\ge \lambda]
		\le 2\cdot \exp (-\frac{m\lambda^2}{O(1/d\epsilon_2^2)+\lambda \cdot O(1/\epsilon_2)}).
		\end{align*}
		By the union bound, there exists $\lambda=O(\frac{\sqrt{\log d/\beta}}{\epsilon_2 \sqrt{md}})$ such that $\max_{j\in[1,d]}|\tilde{s}_j-X_j|<\lambda$ holds with at least $1-\beta$ probability.
		
		\section{Pseudocode}
		\begin{algorithm}[ht]
			\caption{Exponential Private \textbf{SelectOracle} (EXP)}
			\begin{algorithmic}[1]\label{exp}
				\REQUIRE $r^t, \epsilon_1$
				\STATE initialize $sum\leftarrow \sum_{j=1}^d \exp (\frac{\epsilon_1\cdot j}{d-1})$
				\STATE sort $|r^t|$ in ascending order 
				\FOR {$j\in [d]$}
				\STATE $z_j \leftarrow $ the rank of dimension $j$
				\STATE $p_j \leftarrow \frac{exp(\frac{\epsilon_1 \cdot z_j}{d-1})}{sum}$
				\ENDFOR
				\STATE sample an index $j\in [d]$ with probability $p_j$
				\RETURN $j$
			\end{algorithmic}
		\end{algorithm}
		\begin{algorithm}[ht]
			\caption{\textbf{SelectOracle}-Perturbed Encoding (PE)}
			\begin{algorithmic}[1]\label{de}
				\REQUIRE $r^{t}, k, \epsilon_1$
				\STATE initialize $z\leftarrow \{0\}^d, p=\frac{e^{\epsilon_1}}{e^{\epsilon_1} + 1}$
				\STATE sort $|r^t|$ and for $j\in[d]$, set $z_j \leftarrow 1$ if $|r_j^t| \in$ Top-$k$
				\FOR{$j \in [d]$}
				\IF{$z_j = 1$}
				\STATE $\acute{z}_j \leftarrow \begin{cases}
				1, & w.p. ~~p\cr
				0, & w.p. ~~1-p
				\end{cases}$
				\ELSE
				\STATE $\acute{z}_j \leftarrow \begin{cases}
				1, & w.p. ~~1-p\cr
				0, & w.p. ~~p
				\end{cases}$
				\ENDIF
				\ENDFOR
				\STATE perturbed Top-$k$ index list $\mathbb{S}=\{j|\acute{z}_j=1, j\in[d]\}$
				\IF{$\mathbb{S}$ is empty}
				\RETURN $\bot$
				\ELSE
				\STATE sample a dimension $j$ from $\mathbb{S}$
				\RETURN $j$
				\ENDIF
			\end{algorithmic}
		\end{algorithm}
	\vspace{-40pt}
		\begin{algorithm}[ht]
			\caption{\textbf{SelectOracle}-Perturbed Sampling (PS)}
			\begin{algorithmic}[1]\label{bin}
				\REQUIRE $r^{t}, k, \epsilon_1$
				\STATE initialize $z\leftarrow \{0\}^d$
				\STATE sort $|r^t|$ and for $j\in [d]$, set $z_j\leftarrow 1$ if $|r_j^t| \in$ Top-$k$
				\STATE sample $x$ uniformly at random from $[0,1]$
				\IF {$x<\frac{e^{\epsilon_1}\cdot k}{d-k+e^{\epsilon_1}\cdot k}$}
				\STATE randomly sample an index $j\in \{j\in[d]|z_j=1\}$
				\ELSE
				\STATE randomly sample an index $j\in\{j\in[d]|z_j=0\}$
				\ENDIF
				\RETURN $j$
			\end{algorithmic}
		\end{algorithm}
	\end{subappendices}
\end{document}